\documentclass[runningheads]{llncs}
\usepackage[T1]{fontenc}
\newcommand{\corrauth}{\textsuperscript{(\Letter)}}
\usepackage{graphicx}
\usepackage{color}

\usepackage{pdfpages}

\usepackage{multirow}
\usepackage{cite}
\usepackage[misc]{ifsym}
\begin{document}
\title{Deep Learning-based Facial Appearance Simulation Driven by Surgically Planned Craniomaxillofacial Bony Movement}
\titlerunning{Deep Learning-based Facial Appearance Simulation}
%
%
\author{Xi Fang\inst{1*} \and
Daeseung Kim\inst{2*} \and
Xuanang Xu\inst{1} \and
Tianshu Kuang\inst{2} \and
Hannah H. Deng\inst{2} \and
Joshua C. Barber\inst{2} \and
Nathan Lampen\inst{1} \and
Jaime Gateno\inst{2} \and
Michael A.K. Liebschner\inst{3} \and
James J. Xia\inst{2}\corrauth \and
Pingkun Yan\inst{1}\corrauth}
\renewcommand{\thefootnote}{\fnsymbol{footnote}}
\footnotetext[1]{X. Fang and D. Kim contributed equally to this paper.}
\footnotetext[2]{J.J. Xia (\email{JXia@houstonmethodist.org}) and P. Yan (\email{yanp2@rpi.edu}) are co-corresponding authors.}
\authorrunning{X.~Fang et al.}
%
\institute{Department of Biomedical Engineering and Center for Biotechnology and Interdisciplinary Studies, Rensselaer Polytechnic Institute, Troy, NY 12180, USA
\and 
Department of Oral and Maxillofacial Surgery, Houston Methodist Research Institute, Houston, TX 77030, USA
\and 
Department of Neurosurgery, Baylor College of Medicine, Houston, TX 77030, USA}
\maketitle              
\begin{abstract}
Simulating facial appearance change following bony movement is a critical step in orthognathic surgical planning for patients with jaw deformities. Conventional biomechanics-based methods such as the finite-element method (FEM) are labor intensive and computationally inefficient. Deep learning-based approaches can be promising alternatives due to their high computational efficiency and strong modeling capability. However, the existing deep learning-based method ignores the physical correspondence between facial soft tissue and bony segments and thus is significantly less accurate compared to FEM. In this work, we propose an Attentive Correspondence assisted Movement Transformation network (ACMT-Net) to estimate the facial appearance by transforming the bony movement to facial soft tissue through a point-to-point attentive correspondence matrix. Experimental results on patients with jaw deformity show that our proposed method can achieve comparable facial change prediction accuracy compared with the state-of-the-art FEM-based approach with significantly improved computational efficiency. 

\end{abstract}
\section{Introduction}
Orthognathic surgery is a bony surgical procedure (called “osteotomy”) to correct jaw deformities. During orthognathic surgery, the maxilla and the mandible are osteotomized into multiple segments, which are then individually moved to a desired (normalized) position. While orthognathic surgery does not directly operate on facial soft tissue, the facial appearance automatically changes following the bony movement \cite{shafi2013accuracy}. Surgeons now can accurately plan the bony movement using computer-aided surgical simulation (CASS) technology in their daily practice \cite{xia2009new}. However, accurate and efficient prediction of the facial appearance change following bony movement is still a challenging task due to the complicated nonlinear relationship between facial soft tissues and underlying bones \cite{kim2021novel}.

Finite-element method (FEM) is currently acknowledged as the most physically-relevant and accurate method for facial change prediction. However, despite the efforts to accelerate FEM \cite{faure2012sofa,johnsen2015niftysim}, FEM is still time-consuming and labor-intensive because it requires heavy computation and manual mesh modeling to achieve clinically acceptable accuracy \cite{kim2021novel}. In addition, surgical planning for orthognathic surgery often requires multiple times of revisions to achieve ideal surgical outcomes, therefore preventing facial change prediction using FEM from being adopted in daily clinical setting \cite{lampen2022deep}.

Deep learning-based approaches have been recently proposed to automate and accelerate the surgical simulation. 
Li et al. \cite{li2020malocclusion} proposed a spatial transformer network based on the PointNet \cite{qi2017pointnet} to predict tooth displacement for malocclusion treatment planning. Xiao et al. \cite{xiao2021self} developed a self-supervised deep learning framework to estimate normalized facial bony models to guide orthognathic surgical planning. 
However, these studies are not applicable to facial change simulation because they only allows single point set as input whereas facial change simulations require two point sets, i.e., bony and facial surface points. Especially, in-depth modeling of the correlation between bony and facial surfaces is the key factor for accurate facial change prediction using deep learning technology. Ma et al. \cite{ma2021deep} proposed a facial appearance change simulation network, FC-Net, that embedded the bony-facial relationship into facial appearance simulation. FC-Net takes both bony and facial point sets as input to jointly infer the facial change following the bony movement. However, instead of explicitly establishing spatial correspondence between bony and facial point sets, the movement vectors of all bony segments in FC-Net are represented by a single global feature vector. Such a global feature vector ignores local spatial correspondence and may lead to compromised accuracy in facial change simulation.

In this study, we hypothesize that establishing point-to-point correspondence between the bony and facial point sets can accurately transfer the bony movement to the facial points and in turn significantly improve the postoperative facial change prediction. To test our hypothesis, we propose an Attentive Correspondence assisted Movement Transformation network (ACMT-Net) that equipped with a novel cross point-set attention (CPSA) module to explicitly model the spatial correspondence between facial soft tissue and bony segments by computing a point-to-point attentive correspondence matrix between the two point sets. Specifically, we first utilize a pair of PointNet++ networks \cite{qi2017pointnet++} to extract the features from the input bony and facial point sets, respectively. Then, the extracted features are fed to the CPSA module to estimate the point-to-point correspondence between each bony-facial point pair. Finally, the estimated attention matrix is used to transfer the bony movement to the preoperative facial surface to simulate postoperative facial change.

The contributions of our work are two-fold. 1) From the technical perspective, an ACMT-Net with a novel CPSA module is developed to estimate the change of one point set driven by the movement of another point set. The network leverages the local movement vector information by explicitly establishing the spatial correspondence between two point sets 2) From the clinical perspective, the proposed ACMT-Net can achieve a comparable accuracy of the state-of-the-art FEM simulation method, while substantially reducing computational time during the surgical planning. 

\section{Method}
\begin{figure}[t]
	\centering
	\includegraphics[width = \textwidth] {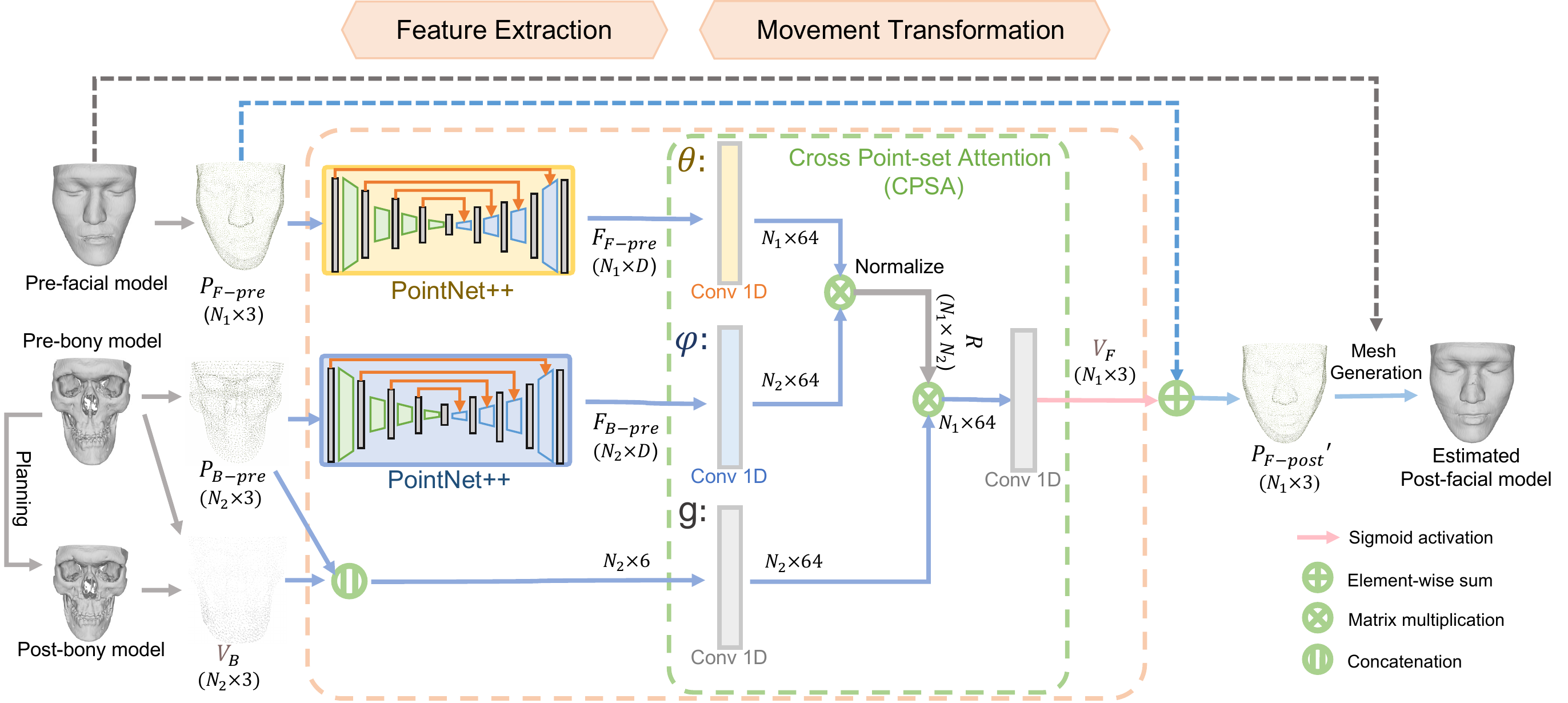}
	\caption{Scheme of the proposed Attentive Correspondence assisted Movement Transformation network (ACMT-Net) for facial change simulation.
	}
	\label{fig:network}
\end{figure}
ACMT-Net predicts postoperative facial model (three-dimensional (3D) surface) based on preoperative facial and bony model and planned postoperative bony model. ACMT-Net is composed of two major components: 1) point-wise feature extraction and 2) point-wise facial movement prediction (Fig.~\ref{fig:network}).
In the first component, pre-facial point set $P_{F-pre}$, pre-bony point set $P_{B-pre}$, and post-bony point set $P_{B-post}$ are subsampled from the pre- and post- facial/bony models for computational efficiency. 
Then the pre-facial/bony point sets ($P_{F-pre}$, $P_{B-pre}$) are fed into a pair of PointNet++ networks to extract semantical and topological features $F_{F-pre}$ and $F_{B-pre}$, respectively. 
In the second component,$F_{F-pre}$ and $F_{B-pre}$ are fed into the CPSA module to estimate the point-to-point correspondence between facial and bony points.
Sequentially, the estimated point-to-point correspondence is combined with pre bony points $P_{B-pre}$ and bony points movement $V_B$ to estimate point-wise facial movement $V_F$ (i.e., the displacement from the pre-facial points $P_{F-pre}$ to the predicted post-facial points $P_{F-post}\prime$). Finally, the estimated point-wise facial movement is added to the pre-facial model to generate the predicted post-facial model. Both are described below in detail.
\subsection{Point-wise feature extraction}
Point-wise facial/bony features are extracted from preoperative facial/bony point sets using PointNet++ networks that are modified by removing the classification layer. The extracted features contain topological and semantic information so that they can be further used to calculate spatial correspondence between facial and bony points. 
\begin{figure}[t]
	\centering
	\includegraphics[width = \textwidth] {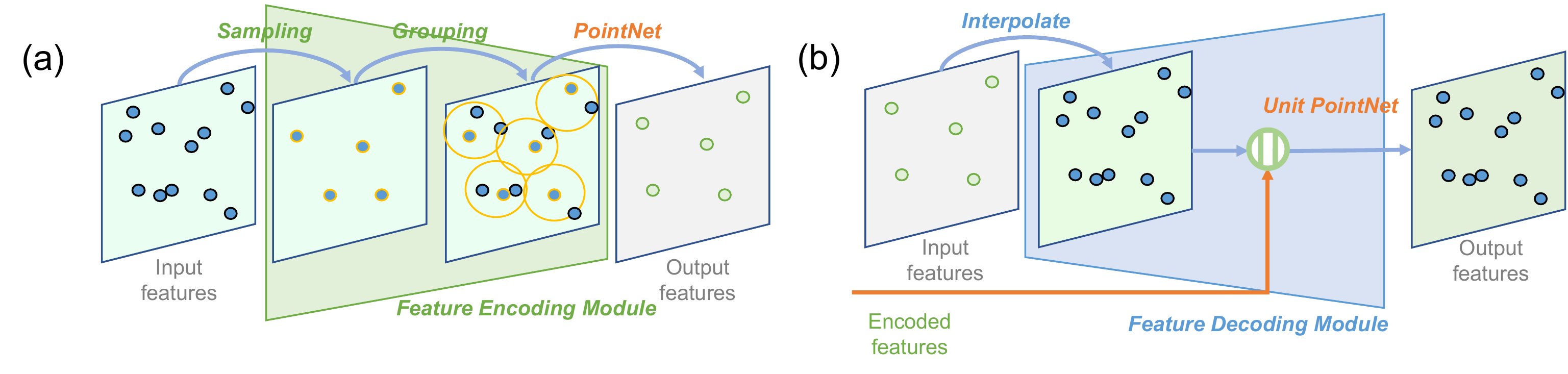}
	\caption{Details of (a) feature encoding module and (b) feature decoding module of PointNet++.
	}
	\label{fig:pointnet}
\end{figure}
Each modified PointNet++ network is composed of a group of feature encoding and decoding modules as shown in Fig.~\ref{fig:pointnet}. In each feature encoding module, a subset of points is firstly down-sampled from the input points using iterative farthest point sampling (FPS). Then each sampled point and its neighboring points (with a searching radius of 0.1) are grouped together and fed into a PointNet to abstract the feature into higher level representations. The PointNet \cite{qi2017pointnet}, composed of a shared MLP layer and a max-pooling layer, captures high-level representations from sets of local points.  In each feature decoding module, the point-wise features are interpolated to the same number of points in corresponding encoding module. The features are concatenated with corresponding features from the encoding module, then a unit PointNet layer extracts the point-wise features. Unit PointNet layer, which is similar to PointNet, extracts each point’s feature vector without sampling and grouping.
\subsection{Attentive correspondence assisted movement transformation}
Attention mechanism can capture the semantical dependencies between two feature vectors at different positions (e.g., space, time) \cite{vaswani2017attention,wang2018non}. Unlike most previous approaches that aim to enhance the feature representation of a position in a sequence (e.g., sentence, image patch) by attending to all positions, our CPSA module learns point-to-point correspondence between positions of two sequences (i.e. point sets) for movement transformation. 

In this step, $F_{F-pre}$, $F_{B-pre}$ and the concatenated pre-bony points and point-wise bony movement vectors $[P_{B-pre}, V_B]$ are fed into a CPSA module to predict the facial movement vectors $V_F$. Firstly, two 1D convolutional layers, $\theta$ and $\varphi$ , are used to map the extracted features of facial and bony point sets to the same embedding space. Then a dot product similarity is computed between the facial feature embedding and bony feature embedding, which represents the relationship, i.e., affinity, between each pair of bony and facial points. 
\begin{equation}
    {f(P_{F-pre},\ P_{B-pre})\ =\ {\theta(F}_{F-pre})}^T{\varphi(F}_{B-pre})
\end{equation}
 The correlation matrix $f(P_{F-pre}$, $P_{B-pre})$ is normalized by the number of bony points sampled from $P_{B-pre}$. The normalized relation matrix $R$ models the attentive point-to-point correspondence between bony points and facial points. 
The movement of facial points is estimated based on the movement of bony points by exploiting the correspondence between bony and facial points. 
Due to the non-linear relationship between facial change and bony movement, we use a 1D convolutional layer, $g$, to compute the representation of point-wise bony movement vectors based on $[P_{B-pre}$, $V_B]$, 
\begin{equation}
    F_{V_B}\ ={g([P}_{B-pre}, V_B])
\end{equation}
Then $R$ is utilized to estimate the facial movement features $F_{V_F}'$ by transforming the bony movement features $F_{V_B}$. Specifically, the estimated facial movement feature ${F_{V_F}^i}'$ of the $i$-th facial point is a normalized summary of all bony movement features $\{{F_{V_B}^j}\}_{\forall j}$ weighted by the corresponding affinities with it:
\begin{equation}
    F_{V_F}^i\ =\ \ \frac{1}{N_2}\sum_{\forall j\ }{{f(P_{F-pre},\ P_{B-pre})}^iF_{V_B}^j},1\leq i \leq N_1,
\end{equation}
where $N_1$ and $N_2$ denote the number of facial/bony points of $P_{F-pre}$ and $P_{B-pre}$, respectively. Sequentially, another 1D convolutional layers reduce the dimension of the facial movement feature to 3. After a sigmoid activation function, the predicted facial movement vector $V_F$ is constrained in [-1,1]. Finally, $V_F$ is added to $P_{F-pre}$ to predict the post-facial points $P_{F-post}’$. 

To train the network, we adopt a hybrid loss function, $Loss=L_{shape} + \alpha L_{density} + \beta L_{LPT}$, to compute the difference between $P_{F-post}’$ and $P_{F-post}$. The loss includes a shape loss $L_{shape}$ \cite{yin2018p2p} to minimize the distance between prediction and target shape, a point density loss $L_{density}$ \cite{yin2018p2p} to measure the similarity between prediction and target shape, and a local-point-transform (LPT) loss $L_{LPT}$ \cite{ma2021deep} to constraint relative movements between one point and its neighbors.  

Since our ACMT-Net processed the point data in a normalized coordinate system, all post-facial points are scaled to the physical coordinate system to generate post-operative faces according to the scale factor of the input points. The movement vectors of all vertices in the pre-facial model are estimated by interpolating the movement vector of 4096 facial points based on the structure information. Predicted post-facial mesh grid is reconstructed by adding the predicted movement vectors to the vertices of pre-facial model.
\section{Experiments and Results}
\subsection{Dataset}
The performance of ACMT-Net was evaluated using 40 sets of patient CT data using 5-fold cross validation. The CT scans were randomly selected from our digital archive of patients who had undergone double-jaw orthognathic surgery (IRB\# Pro00008890). The resolution of the CT images used in this study was $0.49mm\times0.49mm\times1.25mm$.
The segmentation of facial soft tissue and the bones were completed automatically using deep learning-based SkullEngine segmentation tool \cite{liu2021skullengine}, and the surface models were reconstructed using Marching Cube \cite{lorensen1987marching}.
In order to retrospectively recreate the surgical plan that could “achieve” the actual postoperative outcomes, the postoperative facial and bony surface models were registered to the preoperative ones based on surgically unaltered bony volumes, i.e., cranium, and used as a roadmap \cite{ma2021deep}. Virtual osteotomies were then reperformed on preoperative bones. Subsequently, the movement of the bony segments, i.e., the surgical plan to achieve the actual postoperative outcomes, were retrospectively established by manually registering each bony segment to the postoperative roadmap. Finally, the facial appearance change was predicted on the preoperative facial model based on the surgical plan as described below in 3.2. The actual postoperative face served as the ground truth in the evaluation. For efficient training, 4,096 points were down-sampled from each original pre-bony and facial models, respectively. 
\subsection{Implementation and evaluation methods}
We implemented three methods to predict postoperative facial appearance based on the bony movement. They were finite element model with realistic lip sliding effect (FEM-RLSE) [4], FC-Net [6], and our ACMT-Net. While the implementation details of FEM-RLSE and FC-Net can be found in [4,6], the implementation detail of ACMT-Net is described below. 

During the point-wise pre-facial and pre-bony feature extractions, both PointNet++ networks were composed of 4 feature-encoding modules, followed by 4 feature-decoding modules. The output feature dimensions for each module were \{128, 256, 512, 1024, 512, 256, 128, 128\}, respectively. The input point number of the first feature-encoding module was 4,096 and the output point numbers of the extracted point-wise features from each module were \{1024, 512, 256, 64, 256, 512, 1024, 4096\}, respectively. In the CPSA module, the dimension of feature embedding was set to 64. 
We trained the ACMT-Net for 500 epochs with a batch size of 2 using Adam optimizer \cite{kingma2014adam}, after which the trained model was used for evaluation. The learning rate was initialized as 1e-3 and decayed to 1e-4 at discrete intervals during training. And $\alpha$ and $\beta$ were empirically set to be 0.3 and 5, respectively. The models were trained on an NVIDIA DGX-1 deep learning server equipped with eight V100 GPUs. 

The prediction accuracy of ACMT-Net was evaluated quantitatively and qualitatively by comparing it to FEM-RLSE and FC-Net. The quantitative evaluation was to assess the accuracy using the average surface deviation error between the predicted and the actual postoperative facial models \cite{kim2021novel}. 
In addition, using the facial landmarks, we divided the entire face into six regions, including nose, upper lip, lower lip, chin, right cheek and left cheek \cite{kim2021novel}, and calculated the average surface deviation error individually.
Repeated measures Analysis of Variances (ANOVA) was performed to detect whether there was a statistically significant difference among the FEM-RLSE, FC-Net and our ACMT-Net. If significant, post-hoc tests were used to detect the pairwise difference. 
A qualitative evaluation of the upper and lower lips, the most clinically critical region, was carried out by comparing the three predicted lips to the actual postoperative lips. 
One experienced oral surgeon performed the qualitative analysis twice using 2 points Likert scale (1: clinically acceptable; 2: clinically unacceptable). The second analysis was performed a week after the first one to minimize the memory effect. The final decision was made based on the two analysis results. Wilcoxon signed-rank test was performed to detect statistically significant difference. Finally, the time spent on data preparation and computation was recorded.
\subsection{Results}
\begin{figure}[t]
	\centering
	\includegraphics[width = 0.9 \textwidth] {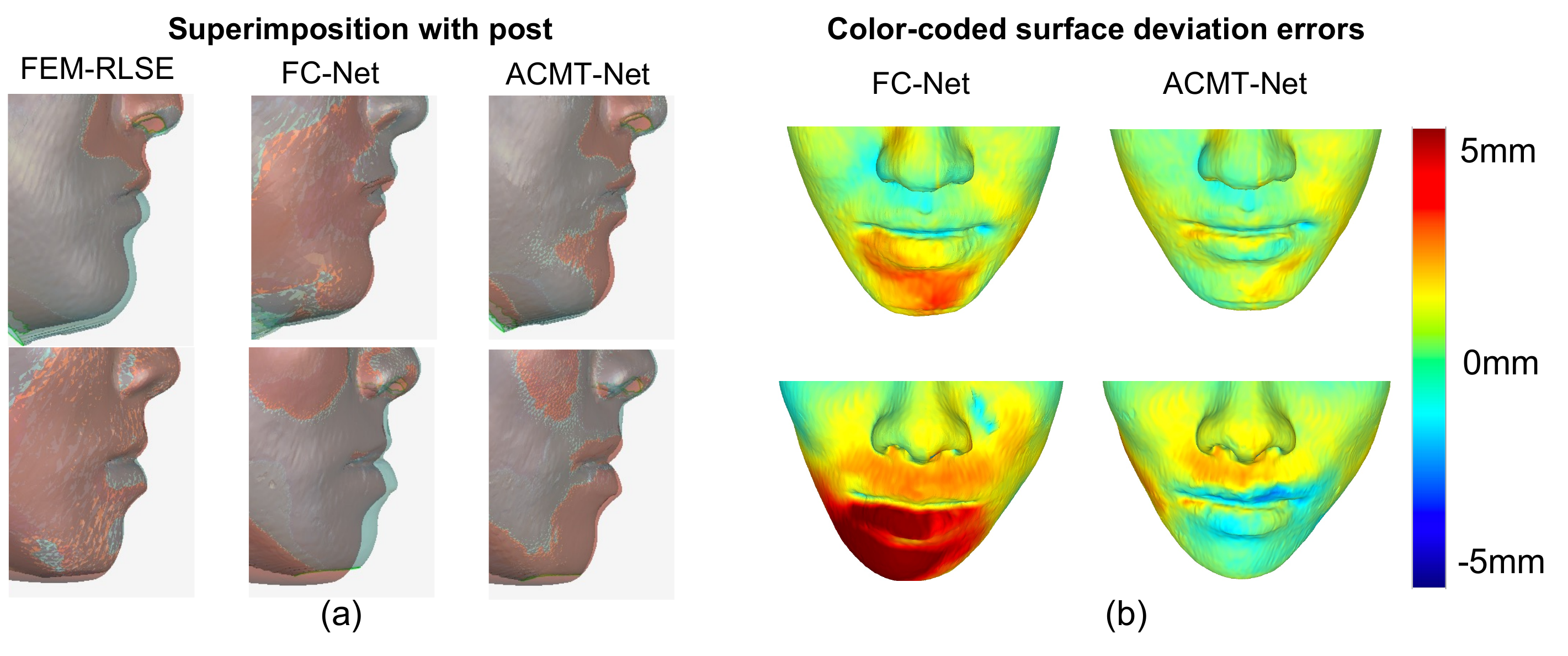}
	\caption{Examples of the simulation results. (a) comparison of the simulated facial models (blue) with ground truth (red) (b) color-coded error map of simulated facial models compared with ground truth.
	}
	\label{fig:result}
\end{figure}
The results of quantitative evaluation are shown in Table 1. The results of repeated measures ANOVA showed there was a statistically significant difference among the three methods (p<0.05), but no significant difference was found among six different regions (p>0.05). The results of post-hoc tests showed that our ACMT-Net statistically significantly outperformed the FC-Net (p<0.05) in all regions. In addition, there was no statistically significance in nose, right cheek, left cheek and entire face (p>0.05) between our ACMT-Net and the state-of-the-art FEM-RLSE method \cite{kim2021novel}. 
The results of the qualitative evaluation showed that among the 40 patients, the simulations of the upper/lower lips were clinically acceptable in 38/38 patients using FEM-RLSE, 32/30 using ACMT-Net, and 24/21 using FC-Net, respectively. 
The results of Wilcoxon signed-rank tests showed that ACMT-Net statistically outperform FC-Net in both lips (p<0.05). In addition, there was no statistically difference between ACMT-Net and FEM-RLSE in the upper lip (p>0.05). 
Fig.~\ref{fig:result} shows two randomly selected patients. It clearly demonstrated that the proposed method was able to successfully predict the facial change following bony movement. Finally it took 70-80 seconds for our ACMT-Net to complete one simulation of facial appearance change. In contrast, it took 30 minutes for FEM-RLSE to completed one simulation.

\begin{table}[t]
\caption{Quantitative evaluation results. Prediction accuracy comparison between ACMT-Net with state-of-the-art FEM-based method and deep learning-based method.}
\scalebox{0.82}{
\begin{tabular}{ |l|c|c|c|c|c|c|c| }
	\hline
	\multirow{2}{6em}{Method} & \multicolumn{7}{|c|}{Surface deviation error (mean ± std) in millimeter}\\
	\cline{2-8}
	& Nose & Upper Lip & Lower Lip & Chin & Right Cheek & Left Cheek & Entire Face \\
	\hline
	FEM-RLSE \cite{kim2021novel} & $0.80\pm0.32$ & $0.96\pm0.58$ & $0.97\pm0.44 $ & $0.74\pm0.54$ & $1.09\pm0.72$ & $1.08\pm0.73$ &$0.97\pm0.48$\\
	FC-Net  \cite{ma2021deep} & $1.11\pm0.60$ & $1.61\pm1.01$ & $1.66\pm0.82$ &	$1.95\pm1.26$ & $1.70\pm0.77$&	$1.59\pm0.94$ &$1.56\pm0.58$\\
	No correspondence & $0.90\pm0.32$ & $1.41\pm0.89$	& $1.84\pm1.05$ & $2.64\pm1.71$ & $1.45\pm0.73$	& $1.59\pm0.73$	& $1.48\pm0.51$\\
	Closest point & $0.81\pm0.30$ & $1.25\pm0.59$ &	$1.34\pm0.66$ &	$1.32\pm0.75$ & $1.12\pm0.63$ & $1.11\pm0.61$ &	$1.10\pm0.38$\\
	ACMT-Net & $0.74\pm0.27$ &	$1.21\pm0.76$ &	$1.33\pm0.72$	& $1.18\pm0.79$ & $1.07\pm0.56$ &	$1.06\pm0.62$	& $1.04\pm0.39$ \\
	\hline
	\end{tabular}}
\end{table}
\subsection{Ablation studies}
We also conducted ablation studies to demonstrate the importance of the spatial correspondence between $F_{F-pre}$ and $F_{B-pre}$ in facial change simulation, and the effectiveness of our CPSA module in modeling such correspondence. Specifically, we considered two ablation models, denoted as ‘No correspondence’ and ‘Closest point’ in Table 1, respectively. For ‘No correspondence’, we solely used the pre-facial point set $P_{F-pre}$ to infer the relation matrix $R$ (by changing the output dimension of $\theta$ from $N_1\times64$ to $N_1\times N_2$, shown in S1). For “Closest point”, we determined the point-to-point correspondence by finding the nearest neighbor point from the pre-facial point set $P_{F-pre}$ to the pre-bony point set $P_{B-pre}$. If the $j$-th bony point was the closest point to the $i$-th facial point, $R(i,j)$ was assigned to 1. Otherwise $R(i,j)$ was assigned to 0. 
As shown in Table 1, compared to the baseline model ‘No correspondence’, the two models with spatial correspondence (‘Closest point’ and our ‘CPSA’) lowered the error from 1.48mm to 1.10mm and 1.04mm, respectively. It suggested that 1) the spatial correspondence was crucial in modeling the physical interaction between facial and bony points; and 2) our design of the CPSA module was more effective than ‘Closest point’ in simulating such correspondence. 

\section{Discussions and Conclusions}
The prediction accuracy evaluation using clinical data proved that ACMT-Net significantly outperformed FC-Net both quantitatively and qualitatively. The results also demonstrated that our ACMT-Net achieved a comparable quantitative accuracy compared with state-of-the-art FEM-RLSE while significantly reducing the prediction time in orthognathic surgical planning. 
The total prediction time is further reduced because our ACMT-Net does not require time-consuming FE mesh generation, thus significantly increases the potentials of using it as a prediction tool in daily clinical practice.
One limitation of our proposed ACMT-Net is the accuracy. 
In FEM-RLSE, 95\% (38/40) of the simulations using FEM-RLSE are clinically acceptable, whereas only about 75\% (32 and 30 of 40) of the simulations using our ACMT-Net are clinically acceptable. 
This may be because that we do not include the anatomical details and the explicit physical interaction during the training due to the efficiency consideration.
In the future, we will improve the simulation accuracy by adaptively sampling the points to include more anatomical details in clinically critical regions and integrate incremental learning strategy into our network.
In particular, our work can be easily extended to other surgical applications that require correspondence mapping between different shapes/point sets \cite{tagliabue2021intra, pfeiffer2020non}.

In conclusion, a deep learning-based framework, ACMT-Net with CPSA module, is successfully implemented to accurately and efficiently predict facial change following bony movements.
The developed network utilizes the spatial correspondence between two associated point sets to transform the movement of one point set to another. Ablation studies also proves the importance and effectiveness of the CPSA module in modeling the spatial correspondence between the associated point sets for movement transformation. 
\\
\\
\noindent \textbf{Acknowledgements.} 
This work was partially supported by NIH under awards R01 DE022676, R01 DE027251 and R01 DE021863.
%

\bibliographystyle{splncs04}
\bibliography{references}


\section*{Supplementary material}

# Supplementary Material
# Deep Learning-based Facial Appearance Simulation Driven by Surgically Planned Craniomaxillofacial Bony Movement

This document provides supplementary material for the paper entitled "Deep Learning-based Facial Appearance Simulation Driven by Surgically Planned Craniomaxillofacial Bony Movement". The supplementary material is organized as follows. In S1, we present the detailed baseline architecture used in Section 3.4. S2 shows a visualization example of the attentive correspondence for interpretation. We find the attentive correspondence learned from ACMT-Net agrees to the area of jaw deformities. S3 presents more color-coded error maps of simulated models compared with the ground truth.

## S1. Baseline model: "No correspondence"

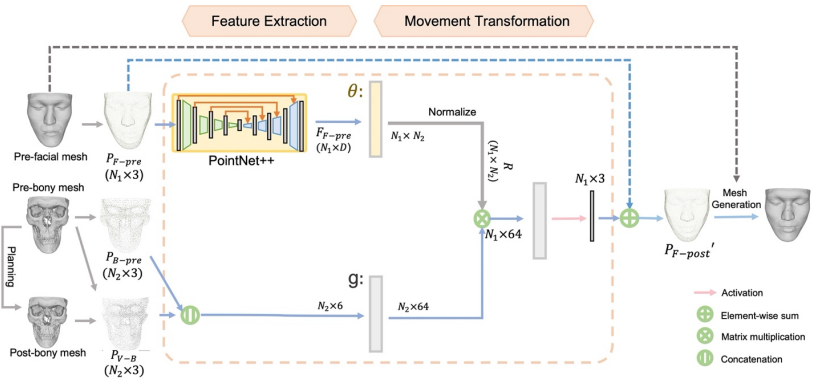

**Fig. 4.** Scheme of the "No correspondence" for facial change simulation. Instead of learning attentive correspondence based on facial and bony points, the architecture solely learns the matrix R to transform bony point-wise movement into facial point-wise movement.

## S2. Visualizations of the attentive correspondence

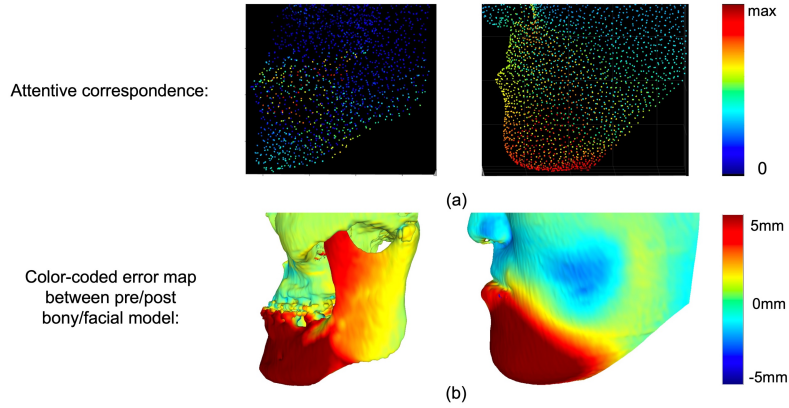

(a)

(b)

**Fig. 5.** (a) Example of the learned attention matrixes. For the example case, we summed up the values along attentive matrix (4096x4096) along horizontal axis (bony points), and vertical axis (facial points), to how the attentive correspondence learned from bony points and projected into facial points. (b) Color-coded error map that shows the distance from pre- bony/facial to post-bony/facial model, respectively. We find the attentive correspondence learned from ACMT-Net agrees to the area of jaw deformities.

## S3. More examples of the color-coded error map of simulated facial models compared with ground truth

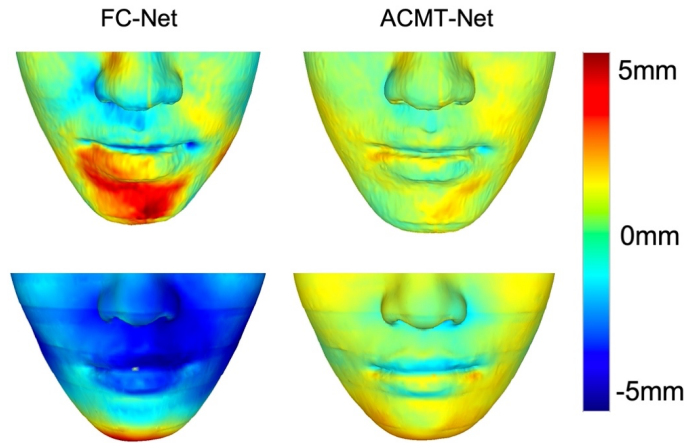

**Fig. 6.** Color-coded error map of simulated facial models compared with ground truth.


\end{document}